\documentclass[letterpaper, 10 pt, conference]{ieeeconf}  

\IEEEoverridecommandlockouts                              

\usepackage{amsmath}
\usepackage{graphicx}
\usepackage{subfigure}
\usepackage{multirow}
\usepackage{booktabs}
\usepackage{threeparttable}
\usepackage{adjustbox}
\usepackage{color}
\definecolor{green}{RGB}{0,255,0}

\usepackage{graphics} 

\title{\LARGE \bf
FIS-Nets: Full-image Supervised Networks for Monocular Depth Estimation
}

\author{Bei Wang$^{1}$ and Jianping An$^{1}$
\thanks{$^{1}$Hikvision Research Institute (HRI), Hangzhou Hikvision Digital Technology Co. Ltd, China.
        {\tt\small wangbei5, anjianping @hikvision.com}}%
}

\begin{document}

\maketitle
\thispagestyle{empty}
\pagestyle{empty}

\begin{abstract}

This paper addresses the importance of full-image supervision for monocular depth estimation. We propose a semi-supervised architecture, which combines both unsupervised framework of using image consistency and supervised framework of dense depth completion. The latter provides full-image depth as supervision for the former. Ego-motion from navigation system is also embedded into the unsupervised framework as output supervision of an inner temporal transform network, making monocular depth estimation better. In the evaluation, we show that our proposed model outperforms other approaches on depth estimation.

\end{abstract}
\section{INTRODUCTION}

Monocular depth estimation (MDE) is essential in scene analysis like autonomous driving or robotics. For an image taken by RGB camera, MDE task aims to predict a depth (i.e. distance to camera) for each pixel in the image. MDE is born as an ill-posed question with “scale uncertainty” since a projection of a single object looks like no difference from before when its size and distance in reality scaled by the same value. Just like what we see a fighting scene in an Ultraman movie, we don’t know whether it is a real building with a super large monster or a small fake building with a man-disguised monster. However, a common sense “the smaller, the further” for a size-fixed object is shared among people. That is how we humans if using only one eye predict roughly an object’s distance from us and understand a scene in real life, with prior knowledge of various object sizes.  

Thanks to the great success and popularity of deep learning methods, depth estimation technique has made much greater progress than ever with using deep networks. Moreover, MDE task is very similar to image semantic segmentation if pixel depth prediction viewed as classification in different depth ranges. As a result, many successful techniques in semantic segmentation task could be directly converted into and applied in depth estimation.

As deep learning are data-driven mechanism, training samples and quality of ground-truth supervision usually becomes key factors in learning process. Neverthless, the problem using existing deep learning methods for MDE is a lack of strong full-image supervision. This work addresses exploring full-image supervision to improve network’s learning ability for MDE.

\section{Related Works}

MDE is a typically difficult issue before the appearance of deep learning methods. Traditional methods take use of image shading or motion to predict object’s structures, like shape-from-shading (SfS) \cite{zhang1999SfS}, structure from motion (SfM) \cite{schonberger2016SfM}\cite{snavely2006SfM}. However, those traditional methods often make several assumptions based on prior knowledge of object description and additional attributes, which somewhat conflict with reality and lack universality. Therefore, those limit their capability of depth estimation in specific scenes or narrow areas. 

Since 2012, deep learning technology has been explosively developed and successfully applied in many fields, including MDE. In the aspect of supervision manner, existing deep learning methods are mainly divided into three streams: supervised, unsupervised and semi-supervised methods. 

\subsection{Supervised Methods}

In supervised methods \cite{eigen2014depth}\cite{liu2015supervised}\cite{fu2018supervised} for depth prediction, ground-truth depth is provided as supervision signal for output of depth prediction network. Simple as it seems, ground-truth depth from LIDAR is usually sparse, which would largely weaken learning capability of depth prediction network. To intensify ground-truth depth density for RGB images, auxiliary sensors like inertial navigation system (imu, GPS, etc.) are used in the process of data capture. LIDAR data from sequential frames are transformed by recorded ego-motion and then fused together in a unified coordinate to produce denser depth for a specific image. However, such a fusion is based on the assumption that all objects in the scene are immobile, otherwise it may induce depth errors if any moving object exists. Even by fusion of sequential LIDAR frames, pixels with ground-truth depth in an image are still rare. In depth prediction task of a public dataset KITTI \cite{geiger2012KITTI}, the greatest density of pixels with ground-truth depth provided in an image is no more than 30\%, which is relatively too sparse to learn. 

\begin{figure}[htbp]
	\centering
	\subfigure[RGB]{
		\includegraphics[width=4.0cm]{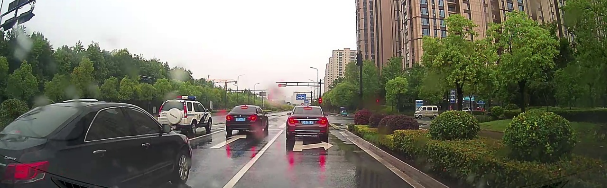}
	}
	\subfigure[Unsupervised]{
		\includegraphics[width=4.0cm]{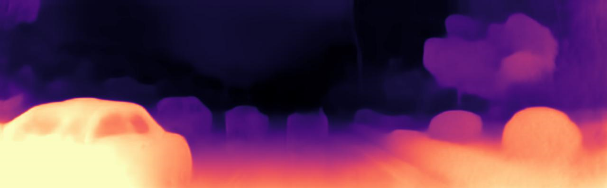}
	}
	\quad
	\subfigure[Sparse-supervised]{
		\includegraphics[width=4.0cm]{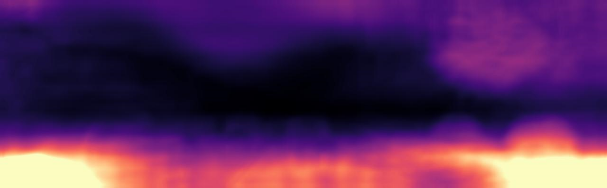}
	}
	\subfigure[Semi-supervised]{
		\includegraphics[width=4.0cm]{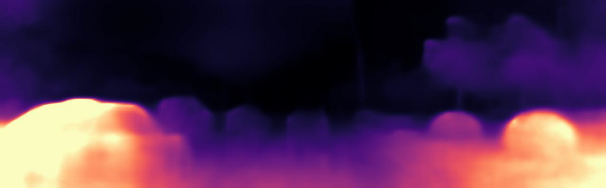}
	}
	\caption{Qualitative visualization of depth prediction from unsupervised, sparse-depth supervised and semi-supervised methods}
	\vspace{-0.5cm}
	\label{figs_3comparison}
\end{figure}

\subsection{Unsupervised Methods}
Unsupervised approaches could be divided into two streams in view of network input: taking temporally continuous images \cite{godard2019unsupervised}\cite{ranjan2019unsupervised}\cite{gordon2019unsupervised}\cite{zhou2017unsupervised}\cite{wang2018unsupervised}\cite{luo2018unsupervised_continuous} or left-and-right-view images \cite{godard2017unsupervised_left_right}\cite{zhan2018unsupervised} as input. The formers predict depth of a single image and inter-frame motion by optimizing consistency losses between image sequences from raw video. The framework often includes a main depth prediction network for a single image and an auxiliary temporal transform network of predicting dynamic inter-frame motion between two successive images (referred to temporal transform network in Fig. \ref{figs_framework1}). A prediction output of temporal transform network is used as a transform matrix to convert 3D points, which are back-projected from image pixels and their predicted depths, into the neighbouring frame. The transformed 3D points are reprojected into a warpped image and make a loss with the original neighbouring image. The loss, called “image reconstruction loss” or “consistency loss”, is then optimized during training time to train the main and the auxiliary networks. In inference time, only the depth prediction network is required, taking a new image as input to predict a depth map. The latter methods are similar but slightly different, constructing a consistency loss between left-and-right-view images. The inner auxiliary network predicts fixed external parameters of left-and-right camera's relative position or simply is omitted if external parameters known.

Compare with sparse-depth-supervised methods, the unsupervised ones has its pros and cons. They calculate and optimize consistency loss between temporally continuous images or left-and-right ones, which is essentially a full-image optimization that cover almost all pixels. As a result, silhouette of objects in image depth prediction look much clearer and more legible than that in sparse-depth-supervised ones, shown in Fig. \ref{figs_3comparison}. However, pixel-wise depth accuracy in an unsupervised way is not guaranteed, if using temporally continuous images from video or left-and-right images without externals known. To be more specific, if a depth map during training scaled in some degree, it would not change consistency loss as long as the translation from temporal transform network or camera's distance scaled in the same degree. In another word, unsupervised approaches hold the property of “scale uncertainty” in nature. A simple but violent way to erase the scale uncertainty is making a statictics about mean scale of ground-truth depth related to model depth prediction for each image at the end of training. Experiment shows the scales basically conform to a Gaussian distribution. A mean scale can be viewed as a fixed scale of ground-truth depth related to model prediction, and be a multiplier of a depth map for final depth prediction. Apparently, such a statistical post-processing is unfriendly and inaccurate due to its unlimited scaling freedom for a single image during training.

\subsection{Semi-supervised Methods}

Semi-supervised approaches \cite{kuznietsov2017semi_supervised} take both sparse-depth-supervised and unsupervised ones discussed above into consideration, aiming to take both their advantages to improve in a combining way.

Our method falls into the third kind, which combines unsupervised framework and depth supervision. However, different from existing semi-supervised methods, this work focuses on building up a depth learning framework with full-image supervision. The full-image supervision includes coarse inter-frame motion from car ego-motion and full-image depths from depth dense completion (DDC) \cite{van2019completion}\cite{hekmatian2019completion}. 

Coarse inter-frame motion is easy-to-get from inertial navigation (like imu) and positioning (like GPS) systems. Supervised by ground-truth ego-motion, the auxiliary temporal transform network in unsupervised framework promotes performance of predicting motion output and thus improves the main network’s capability of predicting a single image’s depth. On the other hand, Supervised by temporal transform, the proposed method combines the nature advantages of both depth-supervised and non-supervised approaches. Therefore, in nature the proposed method biases to the non-supervised one with image re-projection loss in mechanics.

Our contribution is as follows:

\begin{itemize}
	
\item A combined full-image supervised framework is built up to improve learning capability for monocular depth estimation. In the aspect of network architecture, an unsupervised network of depth estimation is combined with a network of dense depth completion to supervise depth prediction. The depth dense completion network gives a full-image depth output, much denser than merged depths from LIDAR frames.

\item Inter-frame motion from navigation system is coarsely calculated and then supervise a temporal transform network in the original unsupervised framework. Inter-frame motion is essentially much more full-image supervision than depth supervision from LIDAR, since dense depth supervision originally comes from fusion of successive LIDAR frames transformed by inter-frame motion. Different from the strong and direct supervision of image depth, such a motion supervision improves depth prediction through supervising temporal transform prediction in a round-about way.
 
\end{itemize}

\section{Overall Architecture}

\begin{figure}[thpb]
	\vspace{-0.5cm}
	\centering
	\includegraphics[scale=0.28]{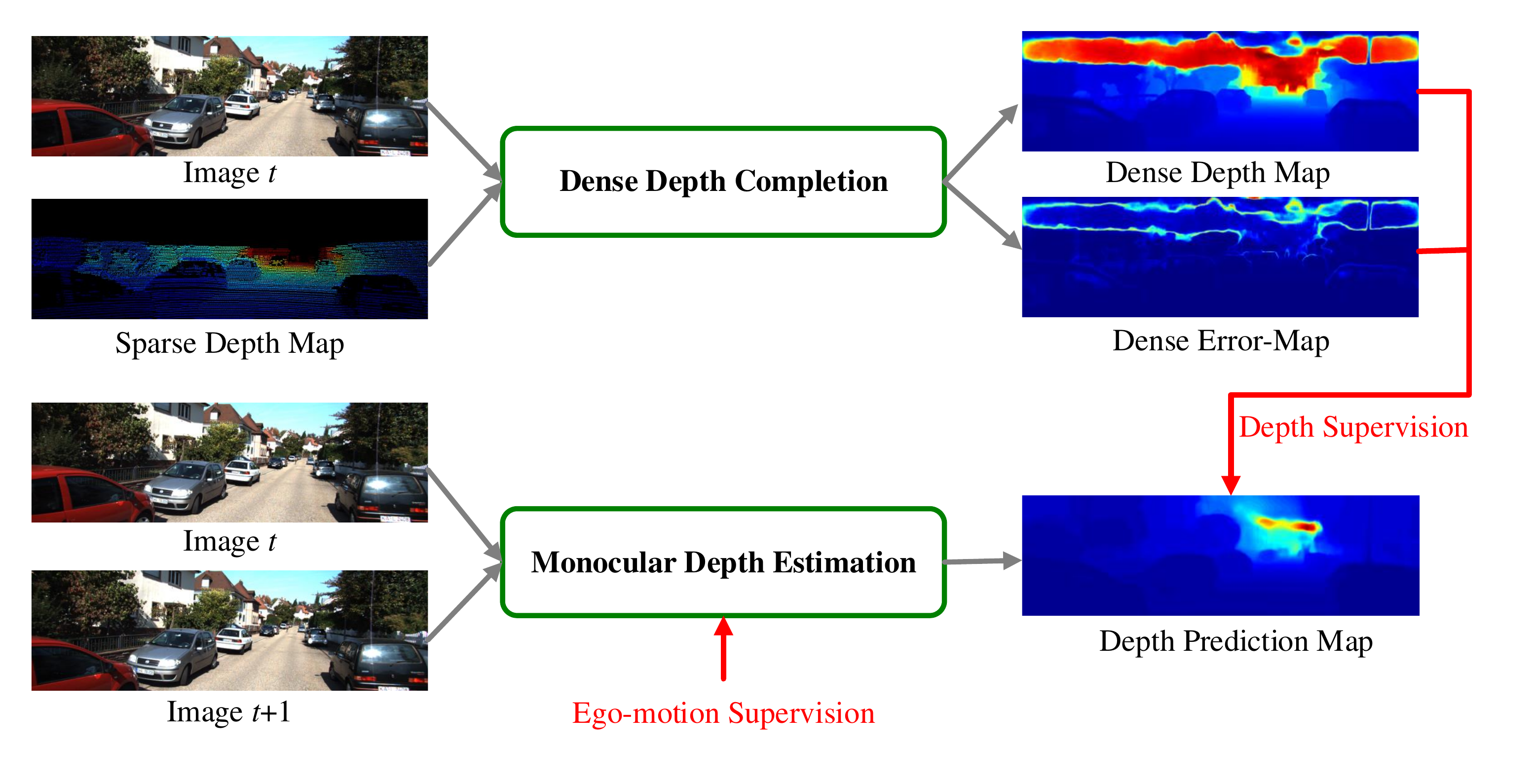}
	\vspace{-0.7cm}
	\caption{Overall architecture of our proposed framework}
	\vspace{-1.0cm}
	\label{figs_architecture}
\end{figure}

The proposed framework, shown in Fig.\ref{figs_architecture}, is a combined framework from two tasks: a main framework for MDE and an auxiliary framework for DDC.

\begin{figure*}[thpb]
	\centering
	\includegraphics[scale=0.45]{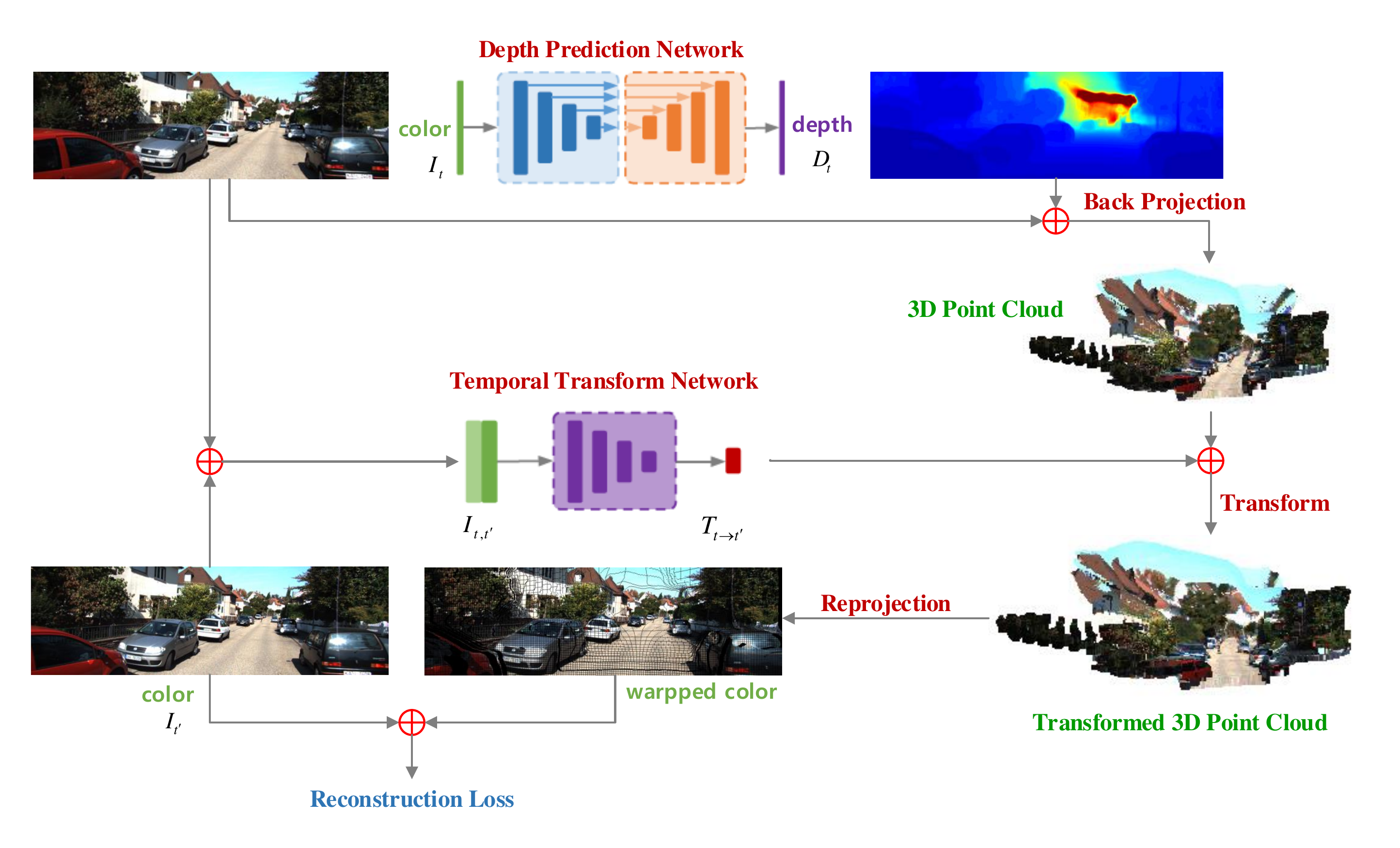}
	\vspace{-0.5cm}
	\caption{Monocular Depth Estimation Framework}
	\vspace{-0.3cm}
	\label{figs_framework1}
	
\end{figure*}
\subsection{Networks}
The main framework is derivated from an unsupervised MDE method, consisting a network of depth prediction for a single image and that of temporal transform prediction for successive image frames, details shown in Fig. \ref{figs_framework1}. The network structure of depth prediction is similar to U-Net, a basic one borrowed from semantic segmentation task, consisted of a top-down encoder and a bottom-up decoder. Depth prediction networks use a classic Resnet18 as an encoder of main depth network and another one as that of temporal transform network. The decoder of the main depth network includes up-convolutions with channels cut half and feature map size doubles until to the original size of input image. Meanwhile, there are some short-cuts linking equal-size layers between the encoder and the decoder of depth network. The decoder of predicting temporal transform simply consists several convolutions to output 6 parameters (3 axis-angles and 3 translations) representing a motion transform matrix. 

We used an auxiliary framework of depth completion proposed in \cite{van2019completion}, which is a RGB-guided and certainty network consisted of global and local branches. The branches are two encoder-decoder networks, taking a RGB image and its corresponding sparse depth map from LIDAR as input and predicting dense depth map as output. Besides, inspired by \cite{hekmatian2019completion}, we also add a header to predict error-map of depth, which can be used to suggest how credible predicted depth is, and also be useful to post-process dense depth maps for generating high-quality depth. The auxiliary framewok aims at dense depth completion to provide full-image depth supervision for the main framework.
\begin{figure*}[thpb]
	\centering
	\includegraphics[scale=0.38]{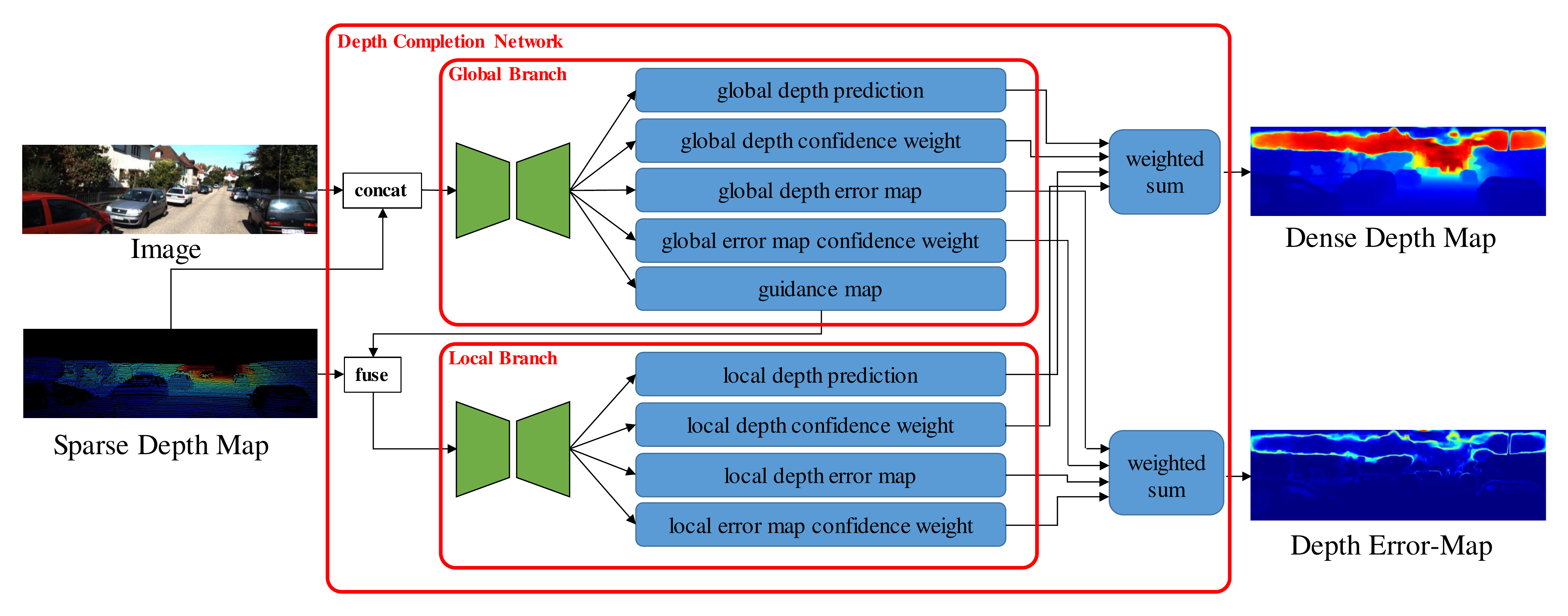}
	\vspace{-0.3cm}
	\caption{Dense Depth Completion Framework}
	\label{figs_framework2}
	\vspace{-0.3cm}
\end{figure*}
The framework for two tasks could be trained simutameouly or separately. In our work, we train the auxiliary dense completion network first and then produce all dense depth image as supervision for training samples in MDE network. Works in literature allege an end-to-end training manner with different task united more elegent and accurate in prediction, however, we found that it is not always the case in actual application. Therefore, we separate these two tasks in independent training to get better accuracy of depth prediction.

\subsection{Full-image Supervision}

The emphasis of this work is full-image supervision from two sources: global ego-motion and dense depth. The ground-truth ego-motion from navigation system supervises outputs of temporal transform network to restrain the freedom of concurrent scaling between predicting depth and inter-frame motion translation. Such ego-motion supervision actually turns an original unsupervising mechanism of the main framework into a supervised one. Small in amount though, ego-motion parameters express strong full-image supervision in essence. On the other hand, dense depth from DDC is induced to directly supervise outputs of depth network, enforcing a stronger restraint on depth prediction.

\subsection{Loss Design}
In our work, depth prediction is simply viewed a pixelwise regression problem, other than a classsification and regression of divided depth ranges. Instead of directly predicting depth, the main depth network outputs a softmax disparity for each pixel, viewed as inverse depth. Similar to \cite{godard2019unsupervised}, we use $d= 1/(ax+b)$ to express conversion between network output disparity x and depth d to basically keep their inherent inverse relations and also limit final prediction in depth range $[d_{min},d_{max}]$. Therefore, if given hyper-parameters $[d_{min},d_{max}]$ ($[0.1,100]$ in our experiments), the final depth prediction is expressed as below:

\begin{equation}\label{depth_conversion}
d = \frac{1}{(\frac{1}{d_{min}}-\frac{1}{d_{max}})x+\frac{1}{d_{max}}}
\end{equation}

As the depth network outputs a softmax disparity $x\in[0,1]$ for each pixel, the final corresponding depth prediction $d$ is limited in $[d_{min},d_{max}]$.

Losses in the proposed framework contains image reconstruction loss $L_{ir}$, disparity smooth loss $L_{ds}$, temporal transform supervision loss $L_{tm}$ and dense depth supervision loss $L_{dd}$. More specific, image reconstruction loss enforces color and texture consistency of temporally continuous frames, by calculating the difference between captured image frame k and a warped image from frame $k+1$ based on predicted depth and motion. Disparity smooth loss constrains an abrupt depth change except regions with big color gradient. Temporal transform supervision loss comes from car-motion while dense depth supervision loss does from forward output of well-trained DDC network for each training sample in MDE. Losses are designed in mathematic formula below:
\begin{equation}\label{losses}
\begin{aligned}
L_{ir} &= \frac{1}{n}\sum_{i}\left\lbrace \alpha\frac{1-SSIM(I_i^k,\tilde{I}_i^k)}{2}+(1-\alpha)\left\|I_i^k-\tilde{I}_i^k \right\| \right\rbrace \\
L_{ds} &= \frac{1}{n}\sum_{i}\left\lbrace \left|\partial_{x}d_i \right|e^{-\left\|\partial_{x}d_i \right\|}+\left|\partial_{y}d_i \right|e^{-\left\|\partial_{y}d_i \right\|}\right\rbrace \\
L_{tm} &= \frac{1}{n}\sum_{i} \left\|T_i-T_i^*\right\|^2,\qquad T_i^*=\left[ R_i^*,   \frac{t_i^*}{s}\right] \\
L_{dd} &= \frac{1}{n}\sum_{i} \frac{(1-err(d_i^*))}{D_i^*}\left\|d_i-D_i^*\right\|^2,\qquad D_i^*=\frac{d_i^*}{s}
\end{aligned}
\end{equation}
Noted that $d_i^*$ here means all-pixels-covered depth prediction from DDC framwork other than sparse ground truth depth provided by KITTI benchmark. In $L_{tm}$ and $L_{dd}$ loss expression above, the same s represents a pre-defined scale to downsize the output target (i.e. ground-truth depth) of depth prediction, aiming to explore the greatest learning capacity of network in an appropriate range. The analysis of hyper-parameter s would go deeper in Section 4.4. As depth has a bigger value in greater distance, loss $L_{dd}$ may make more effort on regression in greater depth. Therefore, a distance-weighted factor $1/D_i^*$ is added in $L_{dd}$ to alleviate the unbalance of regression due to distance difference. In addition, $err(d_i^*)\in[0,1]$ also comes from DDC framework, representing the error probability of pixel-wise depth prediction. Such an error probability could be used to be a loss weight of pixel-wise depth prediction during training.

To dynamically balance the losses above, loss normalization \cite{hekmatian2019completion} is applied during training, so the total loss $L$ is as below:
\begin{equation}\label{total_loss}
L = \frac{L_{ir}}{\left|L_{ir}\right|}+\frac{L_{ds}}{\left|L_{ds}\right|}+\frac{L_{tm}}{\left|L_{tm}\right|}+\frac{L_{d}}{\left|L_{d}\right|}
\end{equation}

\vspace{0.5cm}
\section{Experiments}

\subsection{Evaluation Metrics} Because of scale uncertainty property of unsupervised methods but for a fair comparison with supervised ones, a scale invariant metric, named SILog, is introduced in KITTI benchmark, defined as follows:
\begin{equation}\label{metric_SILog}
D(d,d^*) = \frac{1}{n}\sum_{i} y_i^2 - \frac{1}{n^2}\left(\sum_{i}y_i\right)^2, y_i=\log d_i-\log d_i^*
\end{equation}
where $d_i$ means pixel-wise predicted depth in an image and $d_i^*$ corresonding ground-truth one provided by KITTI benchmark (a slight different with $d_i^*$ in \textit{Loss Design}), $n$ is number of pixels with ground-truth depth in an image. SILog above is specific only for a single image, and an overall SILog metric for evaluating a model is the mean of those for all testing samples. When taking all testing images into consideration, SILog metric becomes:
\begin{equation}\label{metric_SILog2}
D(d,d^*) = \frac{1}{m}\left\lbrace\frac{1}{n}\sum_{i} y_i^2 - \frac{1}{n^2}\left(\sum_{i}y_i\right)^2\right\rbrace
\end{equation}
where $m$ represents the number of testing images for evaluation and $y_i$ defined as in Equation \ref{metric_SILog}.
\linespread{1.5}
\begin{table*}
	\centering
	\caption{Comparison of models using different pre-defined scale $s$}
	\label{tab:scales}
	
	\begin{tabular}{ccccccclccc}
		\hline
		\multirow{2}{*}{Methods}             & \multirow{2}{*}{scale $s$} & \multicolumn{5}{c}{Lower is better}                                                                 &           & \multicolumn{3}{c}{Higher is better}                \\ \cline{3-7} \cline{9-11} 
		&                          & SILog($*10^{-3}$) & AbsRel          & SqRel          & RMSE           & RMSE log        &           & $\delta<1.25$            & $\delta<1.25^2$              & $\delta<1.25^3$              \\ \hline
		\multirow{6}{*}{Supervised by EM + DD} & 1                        & 9.904                         & 0.0618          & 0.258          & 2.869          & 0.0991          &           & 0.9582          & 0.9925          & 0.9981          \\
		& 8                        & 7.859                         & 0.0546          & 0.227          & 2.568          & 0.0885          &           & 0.9701          & 0.9944          & \textbf{0.9982} \\
		& 16                       & 7.626                         & 0.0533          & 0.229          & 2.563          & 0.0871          &           & 0.9717          & 0.9945          & 0.9981          \\
		& 32                       & \textbf{7.309}                & \textbf{0.0513} & \textbf{0.228} & \textbf{2.537} & \textbf{0.0850} & \textbf{} & \textbf{0.9730} & \textbf{0.9946} & 0.9981          \\
		& 64                       & 11.100                        & 0.0665          & 0.289          & 2.712          & 0.1045          &           & 0.9518          & 0.9878          & 0.9962          \\
		& 128                      & 79.011                        & 0.3680          & 2.217          & 4.629          & 0.1045          &           & 0.5542          & 0.7346          & 0.9031          \\ \hline
	\end{tabular}
	\vspace{-0.3cm}
\end{table*}
Other evaluation metrics common used were mean absolute error (MAE), relative absolute error (AbsRel), relative squred error (SqRel), root mean squared error of the depth (RMSE), that of the inverse depth (iRMSE) and so on. These evaluation metrics describes depth errors in absoulte magnitude between predicted and ground-truth depth images.

As the scale uncertainty problem in unsupervised approaches, the metric SILog bridges the gap of difficult comparison between supervised and unsupervised methods. From the observation of Equation (\ref{metric_SILog}), if the predicted depth within an image scaled by a value, the metric SILog never changes. SILog essentially reflects prediction accuracy of relative depth among pixels within an image, other than that of absolute depth. In visualization, a predicted depth image with the lower SILog presents clearer and more explict silhouette of an object, referred into Section 4.3.

\subsection{Dataset Description}
All experiments are conducted on KITTI dataset \cite{geiger2012KITTI} for MDE. Like many previous works, we experiment on the dataset split by Eigen et al.\cite{eigen2014depth}, which contains 23488 training samples from 32 scenes and 697 validating RGB images from remaining 29 scenes. Considering left and right views, the amount of training and validating samples doubles respectively. 

\subsection{Experiments}

Considering the measure limit of LIDAR, $[0.1, 100]$ is taken as ground truth depth range of an image for example. Then the relation between final depth $d$ and predicted softmax disparity $x$ is $d=1/(9.99x+0.01)$. Assuming a Gaussian probability of depth in the range $[0.1, 100]$, median depth values, like 30, 40, 50, are mostly common, and depth extremes, like 0.1 and 100, are rarely seen. Due to the non-linearity of disparity-depth relation, network outputs disparity 0.0023357, 0.0015015, 0.001001 are expected respectively for predicted depth 30, 40 and 50. From the observation above, though the output disparity ranges from 0 to 1, most common depths correspond to very small disparity outputs that almost approximate zero, making a difficulty for network to learn discriminatively. As an experience in deep learning, a network with good learning ability of predicting pixel-wise classification or regression should output discriminative values for pixels with obvisouly different features. To alleviate that problem, a hyper-parameter scale, noted as $s$, is induced to scale down the to-be-learned depth, aiming at scaling up disparity output expectation $x$. In mathematic format, a scaled relation between disparity and depth becomes:

\begin{equation}\label{scaled_depth_conversion}
\begin{aligned}
\frac{d}{s} &= \frac{1}{(\frac{1}{d_{min}}-\frac{1}{d_{max}})x+\frac{1}{d_{max}}}\\
i.e. \quad d &= \frac{s}{(\frac{1}{d_{min}}-\frac{1}{d_{max}})x+\frac{1}{d_{max}}}
\end{aligned}
\end{equation}

\begin{figure*}[htbp]
	\centering
	\subfigure[RGB]{
		\includegraphics[width=5.6cm]{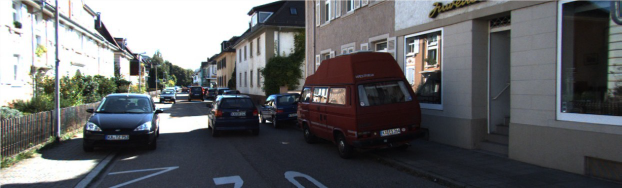}
	}
	\\
	\centering
	\subfigure[$s$ = 1]{
		\includegraphics[width=5.6cm]{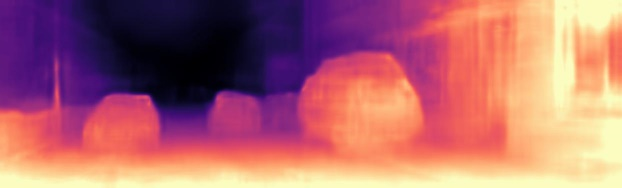}
	}
	\hspace{-0.4cm}
	\subfigure[$s$ = 8]{
		\includegraphics[width=5.6cm]{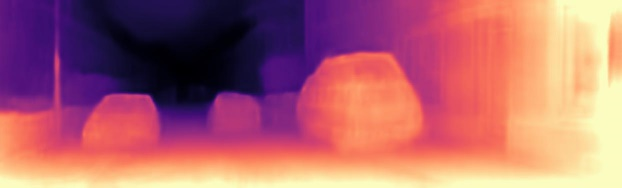}
	}
	\hspace{-0.4cm}
	\subfigure[$s$ = 16]{
		\includegraphics[width=5.6cm]{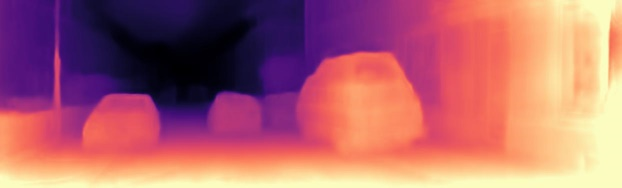}
	}

	\centering
	\subfigure[$s$ = 32]{
		\includegraphics[width=5.6cm]{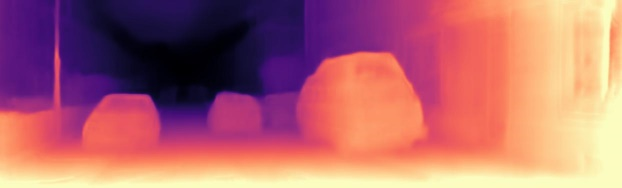}
	}	
	\hspace{-0.4cm}
	\subfigure[$s$ = 64]{
		\includegraphics[width=5.6cm]{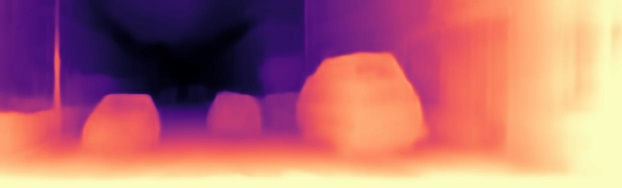}
	}
	\hspace{-0.4cm}
	\subfigure[$s$ = 128]{
		\includegraphics[width=5.6cm]{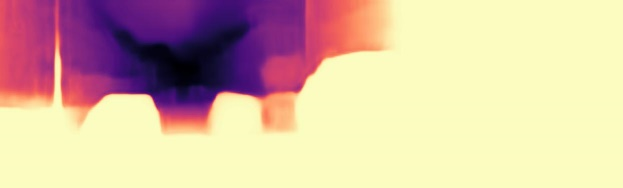}
	}
	\caption{Qualitative results of full-supervised models with different scales for a RGB image}
	\label{figs_scales}
	\vspace{0.3cm}
\end{figure*}

With $s$ induced, network output disparity $x$ would be increased in some degree to get the same depth as before. For example, depth range $[0.1,100]$ and $s=64$ being taken, $x$ should be increased to $0.2125, 0.1591$ and $0.1271$ corresponding to depth 30, 40 and 50 respectively. An increased $x$ would urge the network to learn better with more discriminative disparities of output expectation corresponding to different ground-truth depths. Accordingly, translation in temporal transform would be scaled by the same $s$ to keep the consistency of depth learning target of network. Experiements of depth networks supervised by dense depth (DD) and temporal transform network supervised by ego-motion (EM) with different scales $s$ give out evaluation results in Table \ref{tab:scales} and qualitative visualization of depth prediction for a RGB image in Fig. \ref{figs_scales}.

The evaluation in Table \ref{tab:scales} and qualitative visualization in Fig. \ref{figs_scales} shows that a too small or too large scale could weaken the ability of the depth prediction. A scale of 32 is searched as an optimal hyper-parameter to train the networks.

As the emphasis of this work is full-image supervision, we mainly experiment on the impact of temporal transform (ego-motion) and dense depth supervision on $eigen\_zhou$ dataset splits. To verify the effectiveness of full-image supervision, some experiments on models with or without them together or seperately are conducted, evaluation results shown in Table \ref{tab:U_S}.

\begin{table*}[]
	\centering
	\caption{Comparison of models with or without full-image supervision}
	\label{tab:U_S}
	\begin{threeparttable}
		\begin{tabular}{ccccccccccc}
			\hline
			\multirow{2}{*}{Models} & \multirow{2}{*}{Supervision} & \multicolumn{5}{c}{Lower is better}                                                                &           & \multicolumn{3}{c}{Higher is better}                \\ \cline{3-7} \cline{9-11} 
			&                              & SILog($*10^{-3}$) & AbsRel          & SqRel          & RMSE           & RMSE log       &           & $\delta<1.25$              & $\delta<1.25^2$              & $\delta<1.25^3$              \\ \hline
			U1                      & -                            & 13.27                         & 0.9636          & 14.627         & 18.76          & 3.330          &           & 6e-9            & 1.3e-6          & 4.7e-6          \\
			U2                      & (scaled from U1)             & 12.94                         & 0.1101          & 0.643          & 3.690          & 0.142          &           & 0.8753          & 0.9576          & 0.9952          \\
			U3                      & (scaled from U1)             & 12.92                         & 0.1179          & 0.678          & 3.771          & 0.146          &           & 0.8718          & 0.9601          & 0.9951          \\
			S1                      & ego-motion (EM)              & 12.18                         & 0.0791          & 0.608          & 3.338          & 0.114          &           & 0.9437          & 0.9879          & 0.9951          \\
			S2                      & sparse depth (SD)            & 287.1                         & 0.6770          & 9.161          & 16.00          & 1.367          &           & 0.00535         & 0.0199          & 0.0814          \\
			S3                      & dense depth (DD)             & 8.226                         & 0.0550          & \textbf{0.219} & 2.622          & 0.0904         &           & 0.9664          & 0.9937          & \textbf{0.9982} \\
			S4                      & EM + SD                      & 355.3                         & 0.5594          & 6.004          & 11.388         & 1.106          &           & 0.1621          & 0.2044          & 0.2678          \\
			S5                      & EM + DD                      & \textbf{7.309}                & \textbf{0.0513} & 0.228          & \textbf{2.537} & \textbf{0.085} & \textbf{} & \textbf{0.9730} & \textbf{0.9946} & 0.9981          \\ \hline
		\end{tabular}

		\begin{tablenotes}
			\footnotesize
			\item U represents unsupervised models and S represents supervised methods, including semi-supervised ones.
			\item Hyper-parameter scale s in Supervised Methods (S1-S5) all set 32.
		\end{tablenotes}
	\end{threeparttable}
	\vspace{-0.3cm}
\end{table*}

An unsupervised model U1 without post-processing scaling has a poor performance in each evaluation metric except SILog. Model U2 is scaled by the mean ratio of image ground truth depth median to predition, a scale of $28.027$ based on statistics among all training samples. Model U3 is scaled by ratio of its ground truth depth median to its prediction, a variational ratio for each validate image. In fact, such a scaling in U3 could not be done in testing samples for the ground-truth depth unknown. Thereotically, SILog should keep the same among U1, U2 and U3 since U2 and U3 scaled by different values from U1 and it would not affect SILog as talked before, however because of small magnitude of U1 depth prediction, mathematical accumulated errors make a slight difference among them in SILog.

From the experiements in \ref{tab:U_S}, we could see that either ego-motion supervision or dense depth supervision benefits the prediction accuracy and the combination of them improves it more greatly.

\subsection{Comparison with other methods}
\begin{figure*}[ht] 
	\begin{adjustbox}{
			addcode={
				\hspace{-0.6cm}
				\begin{minipage}{1cm}}{RGB
			\end{minipage}},rotate=90}}
\end{adjustbox}
\hspace{0.06cm}
\begin{minipage}[c]{5.5cm}  
	\includegraphics[width=\textwidth]{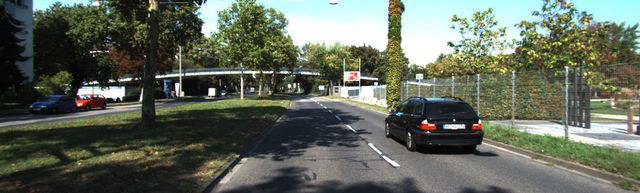} 
\end{minipage} 
\begin{minipage}[c]{5.5cm}  
	\includegraphics[width=\textwidth]{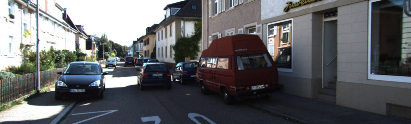} 
\end{minipage} 
\begin{minipage}[c]{5.5cm}  
	\includegraphics[width=\textwidth]{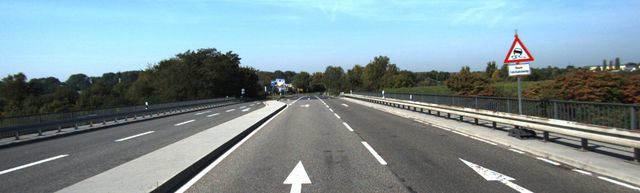} 
\end{minipage} \vspace{0.1cm}
\\
\begin{adjustbox}{
		addcode={
			\hspace{-0.8cm}
			\begin{minipage}{1cm}}{Zhou\cite{zhou2017unsupervised}
		\end{minipage}},rotate=90}}
\end{adjustbox}
\hspace{0.02cm}
\begin{minipage}[c]{5.5cm}  
\includegraphics[width=\textwidth]{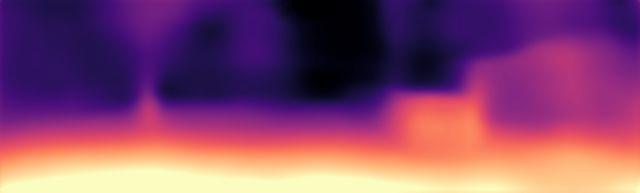} 
\end{minipage} 
\begin{minipage}[c]{5.5cm}  
\includegraphics[width=\textwidth]{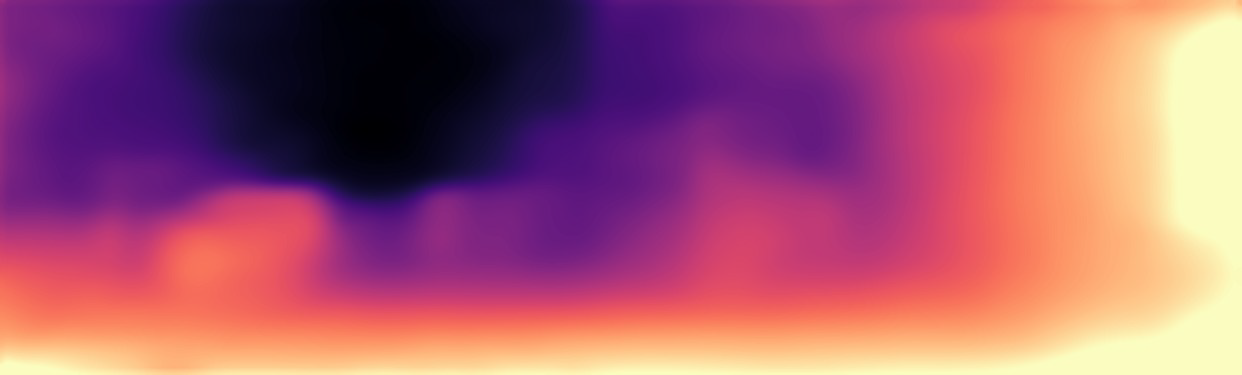} 
\end{minipage} 
\begin{minipage}[c]{5.5cm}  
\includegraphics[width=\textwidth]{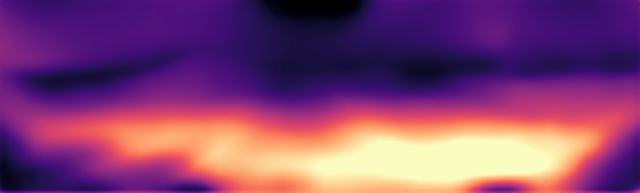} 
\end{minipage}  \vspace{0.1cm}
\\
\begin{adjustbox}{
		addcode={
			\hspace{-0.8cm}
			\begin{minipage}{1cm}}{Zhan\cite{zhan2018unsupervised}
		\end{minipage}},rotate=90}}
\end{adjustbox}
\hspace{0.02cm}
\begin{minipage}[c]{5.5cm}  
\includegraphics[width=\textwidth]{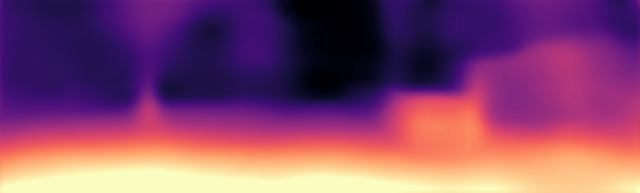} 
\end{minipage} 
\begin{minipage}[c]{5.5cm}  
\includegraphics[width=\textwidth]{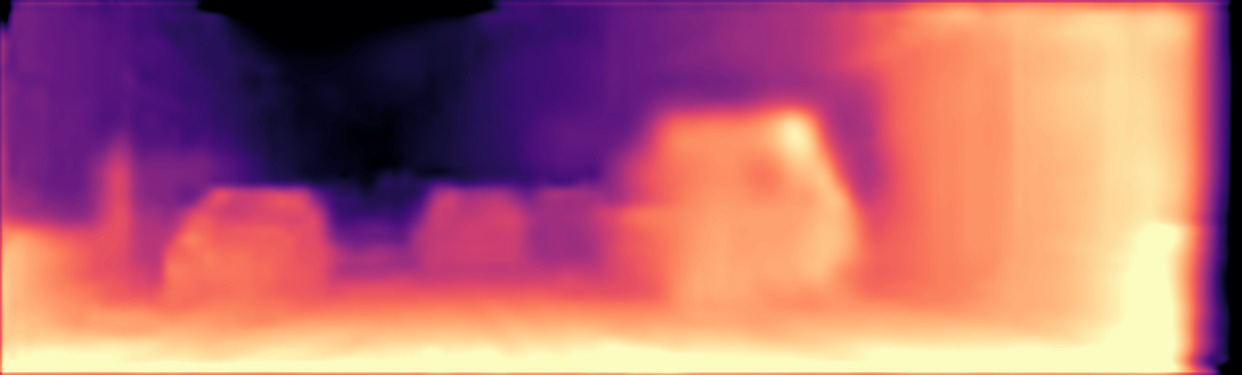} 
\end{minipage} 
\begin{minipage}[c]{5.5cm}  
\includegraphics[width=\textwidth]{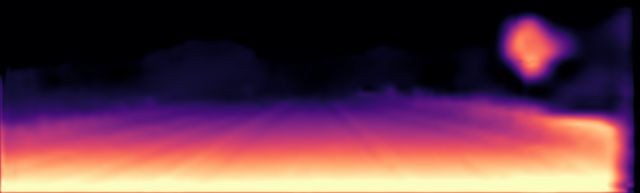} 
\end{minipage} \vspace{0.1cm}
\\
\begin{adjustbox}{
		addcode={
			\hspace{-0.9cm}
			\begin{minipage}{1cm}}{Ranjam\cite{ranjan2019unsupervised}
		\end{minipage}},rotate=90}}
\end{adjustbox}
\hspace{0.02cm}
\begin{minipage}[c]{5.5cm}  
\includegraphics[width=\textwidth]{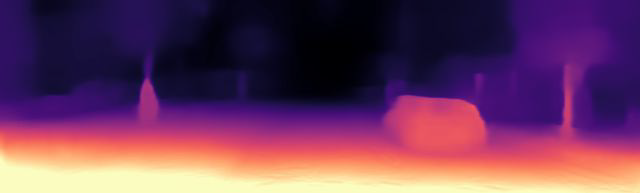} 
\end{minipage} 
\begin{minipage}[c]{5.5cm}  
\includegraphics[width=\textwidth]{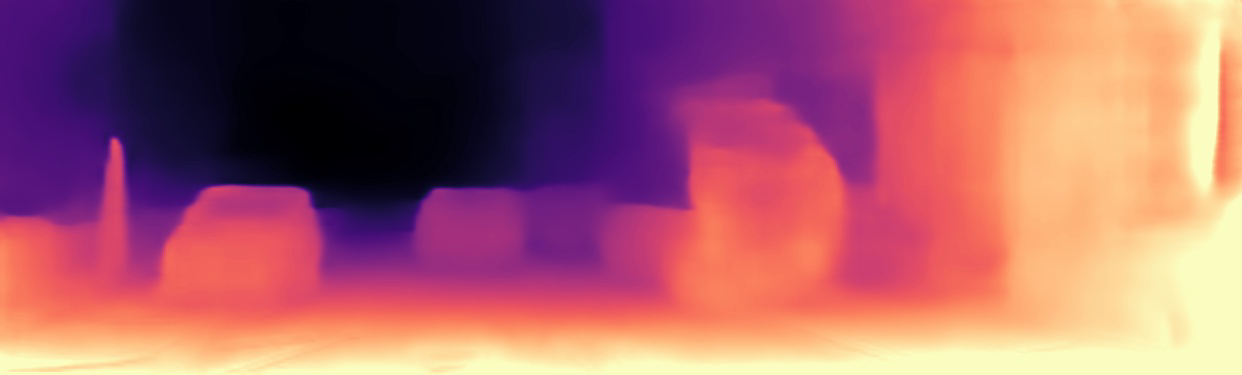} 
\end{minipage} 
\begin{minipage}[c]{5.5cm}  
\includegraphics[width=\textwidth]{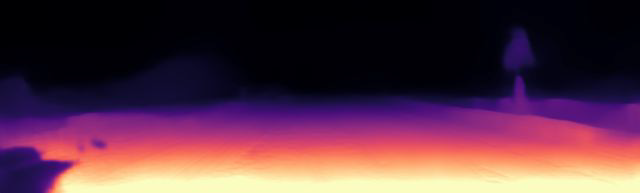} 
\end{minipage} \vspace{0.1cm}
\\
\begin{adjustbox}{
		addcode={
			\hspace{-0.9cm}
			\begin{minipage}{1cm}}{DDVO\cite{wang2018unsupervised}
		\end{minipage}},rotate=90}}
\end{adjustbox}
\hspace{0.02cm}
\begin{minipage}[c]{5.5cm}  
\includegraphics[width=\textwidth]{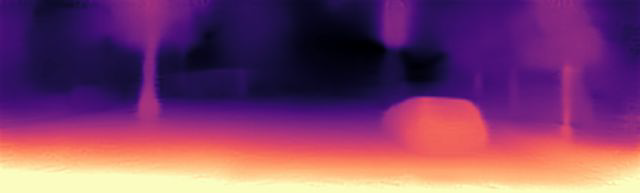} 
\end{minipage} 
\begin{minipage}[c]{5.5cm}  
\includegraphics[width=\textwidth]{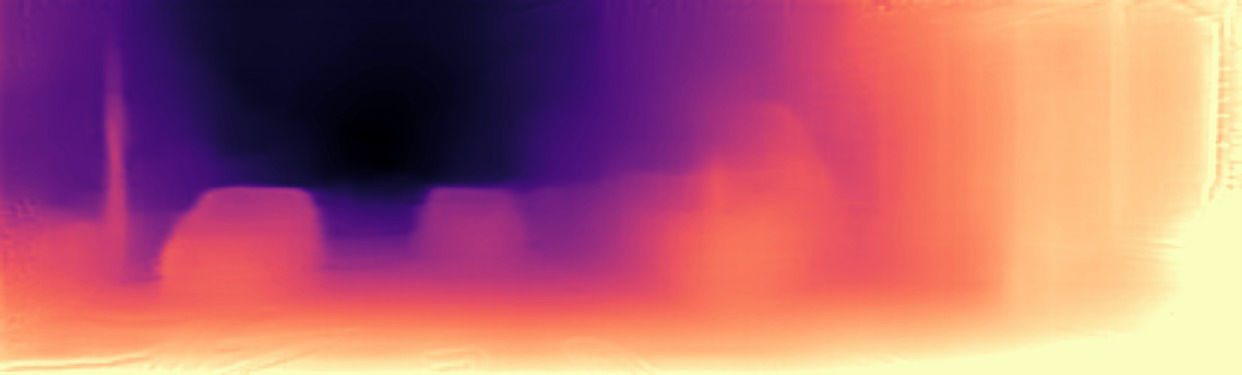} 
\end{minipage} 
\begin{minipage}[c]{5.5cm}  
\includegraphics[width=\textwidth]{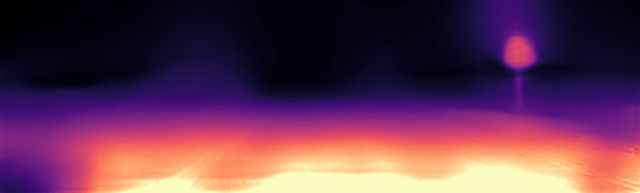} 
\end{minipage} \vspace{0.1cm}
\\
\begin{adjustbox}{
		addcode={
			\hspace{-1.0cm}
			\begin{minipage}{1cm}}{EPC++\cite{luo2018unsupervised_continuous}
		\end{minipage}},rotate=90}}
\end{adjustbox}
\hspace{0.02cm}
\begin{minipage}[c]{5.5cm}  
\includegraphics[width=\textwidth]{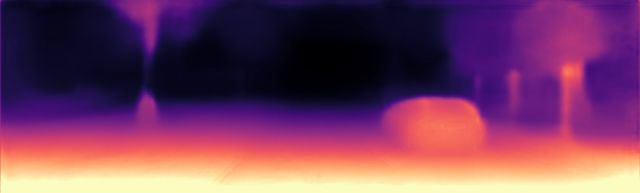} 
\end{minipage} 
\begin{minipage}[c]{5.5cm}  
\includegraphics[width=\textwidth]{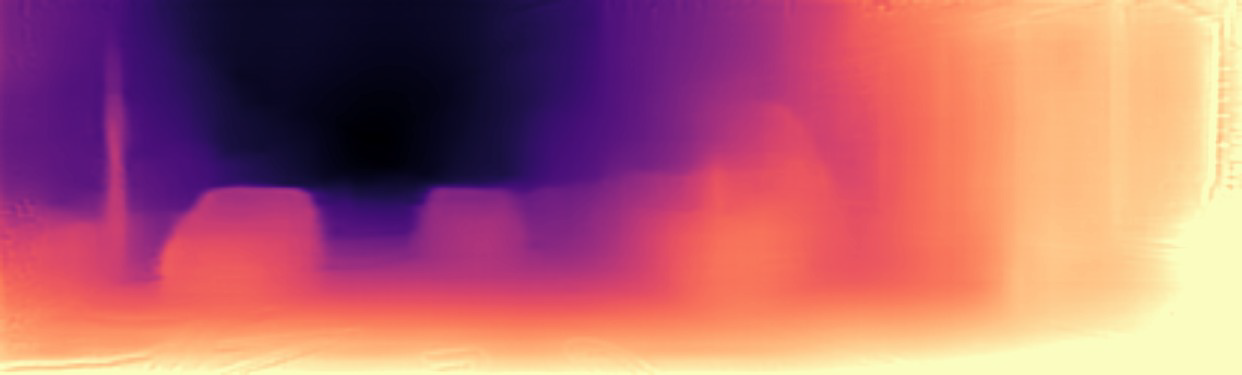} 
\end{minipage} 
\begin{minipage}[c]{5.5cm}  
\includegraphics[width=\textwidth]{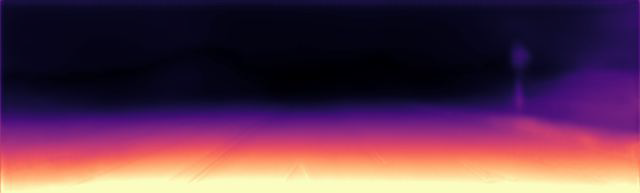} 
\end{minipage} \vspace{0.1cm}
\\
\begin{adjustbox}{
		addcode={
			\hspace{-0.9cm}
			\begin{minipage}{1cm}}{Mono\cite{godard2017unsupervised_left_right}
		\end{minipage}},rotate=90}}
\end{adjustbox}
\hspace{0.02cm}
\begin{minipage}[c]{5.5cm}  
\includegraphics[width=\textwidth]{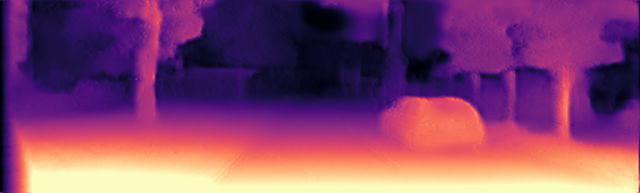} 
\end{minipage} 
\begin{minipage}[c]{5.5cm}  
\includegraphics[width=\textwidth]{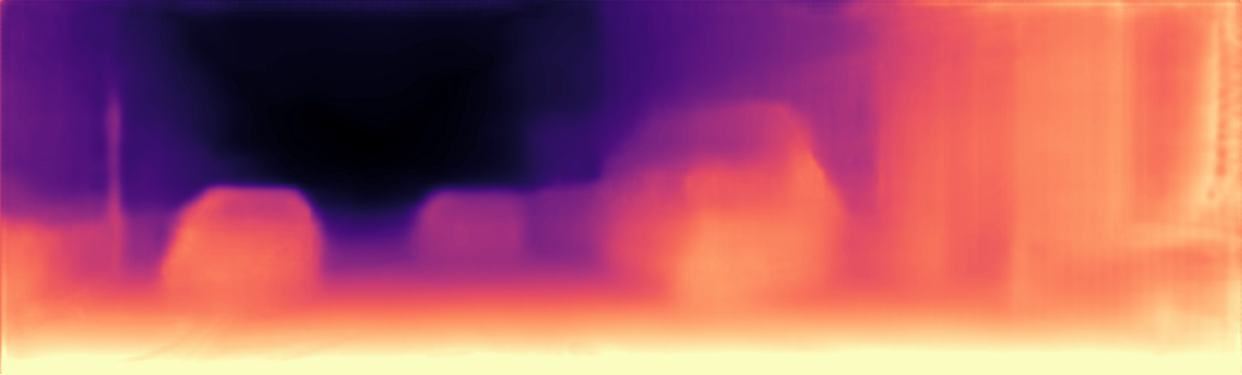} 
\end{minipage} 
\begin{minipage}[c]{5.5cm}  
\includegraphics[width=\textwidth]{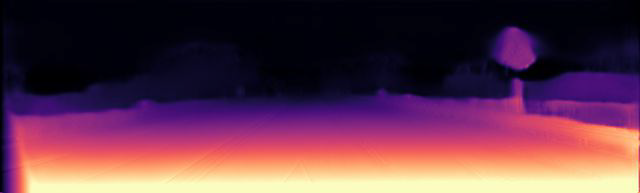} 
\end{minipage} \vspace{0.1cm}
\\
\begin{adjustbox}{
		addcode={
			\hspace{-0.9cm}
			\begin{minipage}{1cm}}{Mono2\cite{godard2019unsupervised}
		\end{minipage}},rotate=90}}
\end{adjustbox}
\hspace{0.02cm}
\begin{minipage}[c]{5.5cm}  
\includegraphics[width=\textwidth]{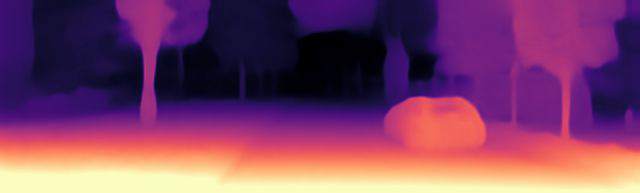} 
\end{minipage} 
\begin{minipage}[c]{5.5cm}  
\includegraphics[width=\textwidth]{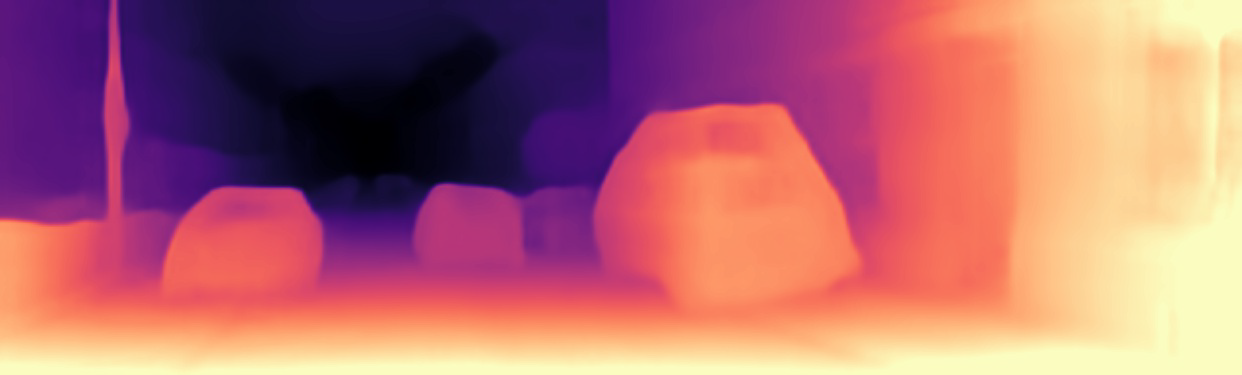} 
\end{minipage} 
\begin{minipage}[c]{5.5cm}  
\includegraphics[width=\textwidth]{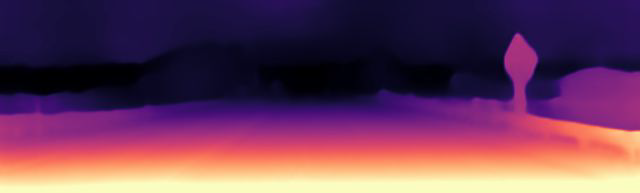} 
\end{minipage} \vspace{0.1cm}
\\
\begin{adjustbox}{
		addcode={
			\hspace{-0.5cm}
			\begin{minipage}{1cm}}{Ours
		\end{minipage}},rotate=90}}
\end{adjustbox}
\hspace{0.06cm}
\begin{minipage}[c]{5.5cm}  
\includegraphics[width=\textwidth]{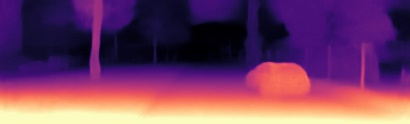} 
\end{minipage} 
\begin{minipage}[c]{5.5cm}  
\includegraphics[width=\textwidth]{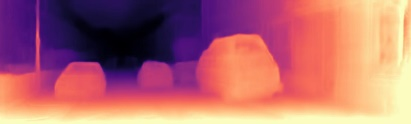} 
\end{minipage} 
\begin{minipage}[c]{5.5cm}  
\includegraphics[width=\textwidth]{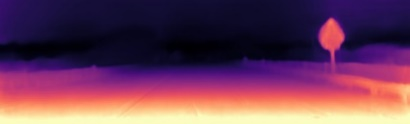} 
\end{minipage} \vspace{0.1cm}
\caption{Comparison of qualitative results on some images with other methods}
\label{figs_other_methods}
\end{figure*}

On the dataset split by Eigen et al\cite{eigen2014depth}, we compare performance of depth prediction with other public methods, listed in \ref{tab:comparsion_methods}. We also compare qualitative visualization of model’s forwarding results with some other methods in Fig. \ref{figs_other_methods}. Comparison of metric evaluation and qualitative visualization with other methods indicates that our model with full-image supervision outperform others in literature. Especially compared with Monodepth \cite{godard2017unsupervised_left_right} and Monodepth2 \cite{godard2019unsupervised} (in the last second and third rows of Fig. \ref{figs_other_methods}), depth prediction maps from our model have clear object silhouettes as good as unsupervised methods and meanwhile outperform them in the evaluation of absolute depth.

\begin{table*}[]
	\centering
	\caption{Comparison of performance on KITTI evaluated on the test split by Eigen et al.}
	\label{tab:comparsion_methods}
	\begin{tabular}{ccccccccc}
		\hline
		\multirow{2}{*}{Methods}      & \multicolumn{4}{c}{Lower is better}                               &           & \multicolumn{3}{c}{Higher is better}             \\ \cline{2-5} \cline{7-9} 
		& AbsRel         & SqRel          & RMSE           & RMSE log       &           & a1             & a2             & a3             \\ \hline
		Eigen et al. Coarse           & 0.361          & 4.826          & 8.102          & 0.377          &           & 0.638          & 0.804          & 0.894          \\
		Eigen et al. Fine             & 0.214          & 1.605          & 6.563          & 0.292          &           & 0.673          & 0.884          & 0.957          \\
		Liu et al. DCNF-FCSP FT*U3    & 0.201          & 1.584          & 6.471          & 0.273          &           & 0.680          & 0.898          & 0.967          \\
		Godard et al.                 & 0.114          & 0.898          & 4.935          & 0.206          &           & 0.861          & 0.949          & 0.976          \\
		Godard et al.                 & 0.116          & 0.918          & 4.872          & 0.193          &           & 0.874          & 0.959          & 0.981          \\
		Kuznietsov et al.             & 0.113          & 0.741          & 4.621          & 0.189          &           & 0.862          & 0.960          & 0.986          \\
		DORN                          & 0.072          & 0.307          & 2.727          & 0.120          &           & 0.932          & 0.984          & 0.994          \\
		Ours (full-image supervision) & \textbf{0.051} & \textbf{0.228} & \textbf{2.537} & \textbf{0.085} & \textbf{} & \textbf{0.973} & \textbf{0.995} & \textbf{0.998} \\ \hline
	\end{tabular}
\end{table*}

\section{Conclusion}

In this work, full-image supervision are emphasisly explored to improve network of depth prediction. Ego-motion is used in depth prediction framework as supervision of temporal transform network's output while dense depth from trained dense completion network used as depth supervision. In network structures, two frameworks of monocular depth estimation and dense depth completion are combined. Extensive experiments demonstrate the full-image supervision from ego-motion and dense depth completion greatly improves performance of depth estimation, outperforming other methods.

\bibliographystyle{IEEEtran}
\bibliography{FISNets}

\begin{thebibliography}{10}
\providecommand{\url}[1]{#1}
\csname url@rmstyle\endcsname
\providecommand{\newblock}{\relax}
\providecommand{\bibinfo}[2]{#2}
\providecommand\BIBentrySTDinterwordspacing{\spaceskip=0pt\relax}
\providecommand\BIBentryALTinterwordstretchfactor{4}
\providecommand\BIBentryALTinterwordspacing{\spaceskip=\fontdimen2\font plus
\BIBentryALTinterwordstretchfactor\fontdimen3\font minus
  \fontdimen4\font\relax}
\providecommand\BIBforeignlanguage[2]{{%
\expandafter\ifx\csname l@#1\endcsname\relax
\typeout{** WARNING: IEEEtran.bst: No hyphenation pattern has been}%
\typeout{** loaded for the language `#1'. Using the pattern for}%
\typeout{** the default language instead.}%
\else
\language=\csname l@#1\endcsname
\fi
#2}}

\bibitem{zhang1999SfS}
R.~Zhang, P.-S. Tsai, J.~E. Cryer, and M.~Shah, ``Shape-from-shading: a
  survey,'' \emph{IEEE transactions on pattern analysis and machine
  intelligence}, vol.~21, no.~8, pp. 690--706, 1999.

\bibitem{schonberger2016SfM}
J.~L. Schonberger and J.-M. Frahm, ``Structure-from-motion revisited,'' in
  \emph{Proceedings of the IEEE Conference on Computer Vision and Pattern
  Recognition}, 2016, pp. 4104--4113.

\bibitem{snavely2006SfM}
N.~Snavely, S.~M. Seitz, and R.~Szeliski, ``Photo tourism: exploring photo
  collections in 3d,'' in \emph{ACM transactions on graphics (TOG)}, vol.~25,
  no.~3.\hskip 1em plus 0.5em minus 0.4em\relax ACM, 2006, pp. 835--846.

\bibitem{eigen2014depth}
D.~Eigen, C.~Puhrsch, and R.~Fergus, ``Depth map prediction from a single image
  using a multi-scale deep network,'' in \emph{Advances in neural information
  processing systems}, 2014, pp. 2366--2374.

\bibitem{liu2015supervised}
F.~Liu, C.~Shen, G.~Lin, and I.~Reid, ``Learning depth from single monocular
  images using deep convolutional neural fields,'' \emph{IEEE transactions on
  pattern analysis and machine intelligence}, vol.~38, no.~10, pp. 2024--2039,
  2015.

\bibitem{fu2018supervised}
H.~Fu, M.~Gong, C.~Wang, K.~Batmanghelich, and D.~Tao, ``Deep ordinal
  regression network for monocular depth estimation,'' in \emph{Proceedings of
  the IEEE Conference on Computer Vision and Pattern Recognition}, 2018, pp.
  2002--2011.

\bibitem{geiger2012KITTI}
A.~Geiger, P.~Lenz, and R.~Urtasun, ``Are we ready for autonomous driving? the
  kitti vision benchmark suite,'' in \emph{2012 IEEE Conference on Computer
  Vision and Pattern Recognition}.\hskip 1em plus 0.5em minus 0.4em\relax IEEE,
  2012, pp. 3354--3361.

\bibitem{godard2019unsupervised}
C.~Godard, O.~Mac~Aodha, M.~Firman, and G.~J. Brostow, ``Digging into
  self-supervised monocular depth estimation,'' in \emph{Proceedings of the
  IEEE International Conference on Computer Vision}, 2019, pp. 3828--3838.

\bibitem{ranjan2019unsupervised}
A.~Ranjan, V.~Jampani, L.~Balles, K.~Kim, D.~Sun, J.~Wulff, and M.~J. Black,
  ``Competitive collaboration: Joint unsupervised learning of depth, camera
  motion, optical flow and motion segmentation,'' in \emph{Proceedings of the
  IEEE Conference on Computer Vision and Pattern Recognition}, 2019, pp.
  12\,240--12\,249.

\bibitem{gordon2019unsupervised}
A.~Gordon, H.~Li, R.~Jonschkowski, and A.~Angelova, ``Depth from videos in the
  wild: Unsupervised monocular depth learning from unknown cameras,''
  \emph{arXiv preprint arXiv:1904.04998}, 2019.

\bibitem{zhou2017unsupervised}
T.~Zhou, M.~Brown, N.~Snavely, and D.~G. Lowe, ``Unsupervised learning of depth
  and ego-motion from video,'' in \emph{Proceedings of the IEEE Conference on
  Computer Vision and Pattern Recognition}, 2017, pp. 1851--1858.

\bibitem{wang2018unsupervised}
C.~Wang, J.~Miguel~Buenaposada, R.~Zhu, and S.~Lucey, ``Learning depth from
  monocular videos using direct methods,'' in \emph{Proceedings of the IEEE
  Conference on Computer Vision and Pattern Recognition}, 2018, pp. 2022--2030.

\bibitem{luo2018unsupervised_continuous}
C.~Luo, Z.~Yang, P.~Wang, Y.~Wang, W.~Xu, R.~Nevatia, and A.~Yuille, ``Every
  pixel counts++: Joint learning of geometry and motion with 3d holistic
  understanding,'' \emph{arXiv preprint arXiv:1810.06125}, 2018.

\bibitem{godard2017unsupervised_left_right}
C.~Godard, O.~Mac~Aodha, and G.~J. Brostow, ``Unsupervised monocular depth
  estimation with left-right consistency,'' in \emph{Proceedings of the IEEE
  Conference on Computer Vision and Pattern Recognition}, 2017, pp. 270--279.

\bibitem{zhan2018unsupervised}
H.~Zhan, R.~Garg, C.~Saroj~Weerasekera, K.~Li, H.~Agarwal, and I.~Reid,
  ``Unsupervised learning of monocular depth estimation and visual odometry
  with deep feature reconstruction,'' in \emph{Proceedings of the IEEE
  Conference on Computer Vision and Pattern Recognition}, 2018, pp. 340--349.

\bibitem{kuznietsov2017semi_supervised}
Y.~Kuznietsov, J.~Stuckler, and B.~Leibe, ``Semi-supervised deep learning for
  monocular depth map prediction,'' in \emph{Proceedings of the IEEE Conference
  on Computer Vision and Pattern Recognition}, 2017, pp. 6647--6655.

\bibitem{van2019completion}
W.~Van~Gansbeke, D.~Neven, B.~De~Brabandere, and L.~Van~Gool, ``Sparse and
  noisy lidar completion with rgb guidance and uncertainty,'' \emph{arXiv
  preprint arXiv:1902.05356}, 2019.

\bibitem{hekmatian2019completion}
H.~Hekmatian, S.~Al-Stouhi, and J.~Jin, ``Conf-net: Predicting depth completion
  error-map forhigh-confidence dense 3d point-cloud,'' \emph{arXiv preprint
  arXiv:1907.10148}, 2019.

\end{thebibliography}
\end{document}